\documentclass[10pt,twocolumn,letterpaper]{article}

\usepackage{wacv}
\usepackage{times}
\usepackage{epsfig}
\usepackage{graphicx}
\usepackage{amsmath}
\usepackage{amssymb}

\usepackage{algorithm}
\usepackage{booktabs}
\usepackage[noend]{algpseudocode}
\usepackage{multirow}

\wacvfinalcopy 

\def\wacvPaperID{212} 

\ifwacvfinal\fi
\begin{document}

\title{Recurrent Autoregressive Networks for Online Multi-Object Tracking}

\author{
Kuan Fang$^{1}$, Yu Xiang$^{2}$, Xiaocheng Li$^{1}$, Silvio Savarese$^{1}$\\
$^{1}$Stanford University, $^{2}$ University of Washington\\
{\tt\small \{kuanfang, chengli1, ssilvio\}@stanford.edu, yuxiang@cs.washington.edu}
}

\maketitle
\ifwacvfinal\thispagestyle{empty}\fi

\begin{abstract}
The main challenge of online multi-object tracking is to reliably associate object trajectories with detections in each video frame based on their tracking history. In this work, we propose the Recurrent Autoregressive Network (RAN), a temporal generative modeling framework to characterize the appearance and motion dynamics of multiple objects over time. The RAN couples an external memory and an internal memory. The external memory explicitly stores previous inputs of each trajectory in a time window, while the internal memory learns to summarize long-term tracking history and associate detections by processing the external memory. We conduct experiments on the MOT 2015 and 2016 datasets to demonstrate the robustness of our tracking method in highly crowded and occluded scenes. Our method achieves top-ranked results on the two benchmarks.
\end{abstract}
\vspace{-6mm}

\section{Introduction}

Tracking multiple objects in videos is an important problem and can be applied to various applications such as visual surveillance, activity analysis, autonomous driving and robot navigation. A common strategy to tackle multi-object tracking is to employ the ``tracking-by-detection'' framework, where the objects are first identified by an object detector in each video frame and then linked into trajectories across video frames. In this case, the core problem is to estimate the characteristics of trajectories over time and thus to determine their associations with new detections.

\begin{figure}
	\centering
	\includegraphics[width = \linewidth]{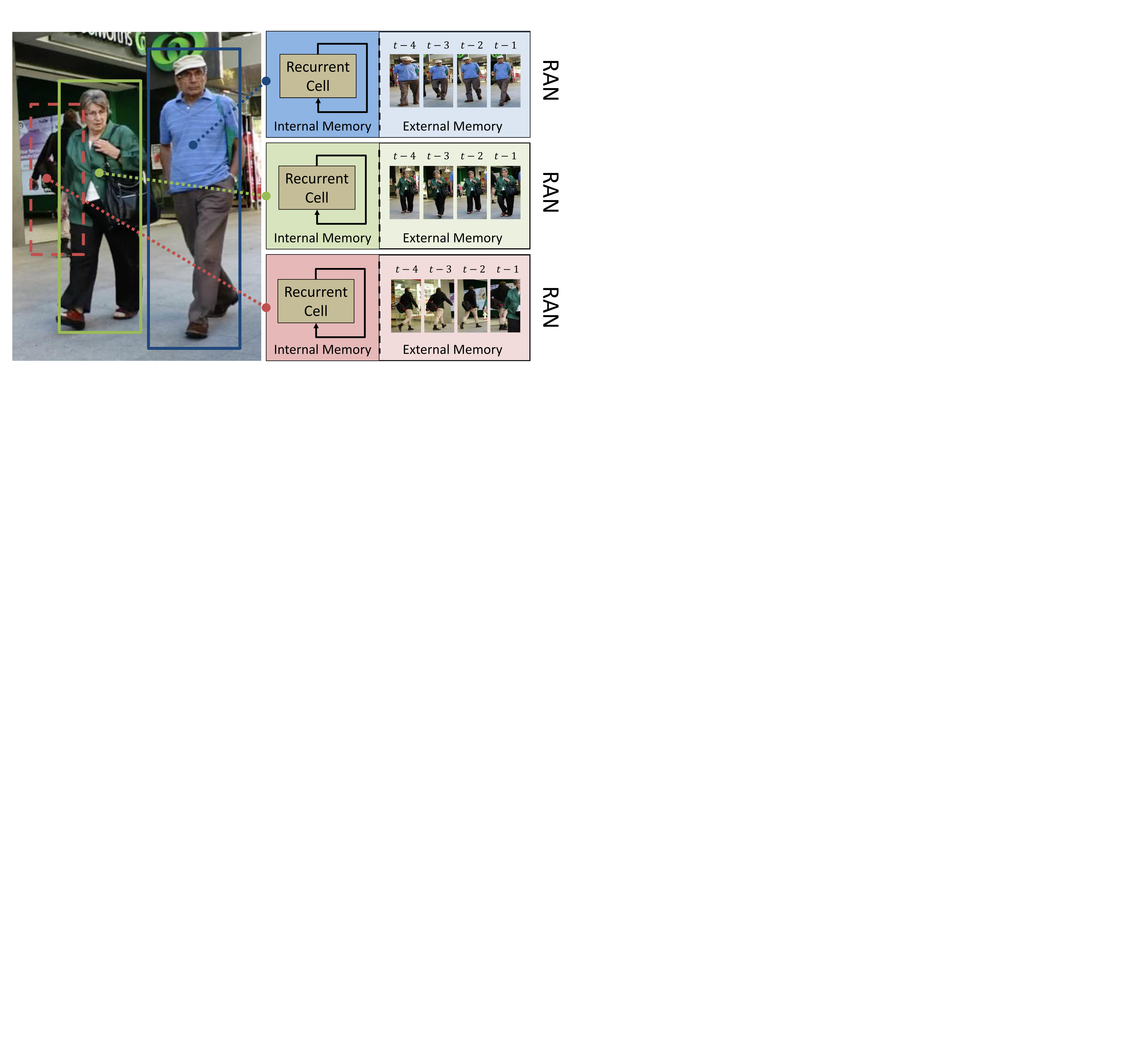}
	\caption{We introduce Recurrent Autoregressive Networks (RANs) for online multiple object tracking. RANs are generative models of object trajectories, which combine external memories of object appearance and motion and internal memories in recurrent cells of the network to facilitate data association in tracking.}
	\label{fig:intro}
    \vspace{-4mm}
\end{figure}

The characteristics of an object trajectory can be depicted by features of the object's appearance and location, which are usually represented by either hand-crafted features (e.g. color histogram, image gradients, optical flow, object coordinates) \cite{butt2013multi, milan2014continuous, choi2015near,xiang2015learning} or neural network extracted features\cite{Kim2015MultipleHT, Ondruska2016DeepTS}. In order to summarize and denoise these features across time, many previous methods \cite{butt2013multi, milan2014continuous, choi2015near,xiang2015learning} store the features in recently tracked frames in templates and use the templates to determine the associations with new detections. Since the templates are constrained in a fixed time window, these models cannot learn the information of long-term history and adjust the estimation accordingly. Recently, there have also been works trying to use recurrent neural networks (RNNs) to extract long-term temporal features for multiple object tracking \cite{Milan2017OnlineMT, Ondruska2016DeepTS}. However, because the feature space of object appearance can be very complicated while the existing multi-object tracking datasets are relatively small, it is hard to train RNNs with enough representation capabilities to memorize and discriminate different objects on existing datasets without overfitting. 

In this work, we propose a recurrent autoregressive network (RAN) framework to learn a generative model for multiple object trajectories. Built upon standard RNN components, the RAN learns to extract long-term sequence history in an internal memory represented by the recurrent hidden layer. To enhance the memorizing power of the model while guaranteeing the generalization capability, we equip our RANs with an external memory as templates directly storing the previous input features. At each time step, the RAN estimates the probability distribution of new detections by using both internal and external memories in an autoregressive manner \cite{Cai1998FunctionalcoeecientRM}. Thus the raw data are memorized by the external memory, while the internal memory focuses on learning to retrieve and process the data from the external memory. Our RAN model can be seen as a variant of memory networks \cite{Sukhbaatar2015EndToEndMN, Graves2014NeuralTM, Xiong2016DynamicMN}. Comparing with traditional memory networks, the RAN constantly updates the memory in a temporal sliding window according to the tracking decisions, instead of using an RNN to control the data reading procedure.

The advantages of RANs are mainly two-folds. First, it enables us to maintain an external memory of the trajectories thus being more robust to occlusions and sudden changes of the targets. Second, the output space of the RAN is a set of parameters that are applied to the templates, which is much easier to train compared with directly estimating a high dimensional feature on smaller tracking datasets. To track multiple objects, we ensemble multiple RANs by attaching one RAN to each target. Furthermore, we design a data association algorithm based on the RAN multi-object tracking framework. The RANs update the hidden states and the templates according to the decisions of the data association algorithm. In this way, they memorize and update the feature representations of the objects in time. Fig. \ref{fig:intro} illustrates an example of the RANs in tracking multiple targets in a video. 

We conducted experiments on the Multiple Object Tracking Benchmark \cite{MOTChallenge2015} to evaluate our RANs for online MOT. From system analysis and comparison with state-of-the-art online multi-object tracking methods, we demonstrate that our RANs are capable of learning a powerful generative model to improve the multi-object tracking performance.

To summarize, our paper has these key contributions:
\begin{itemize}
\item We propose a novel a temporal generative modeling framework: recurrent autoregressive networks (RANs), which couples internal memories (recurrent cells) and external memories (templates in temporal sliding window) to estimate the conditional probability of future sequence. The RAN framework can be potentially applied to various tasks in computer vision and sequential data modeling.
\item We design a multi-object tracking method by associating each object trajectory with an RAN. 
\item Our method outperforms the state-of-the-art online tracking methods on the MOT benchmark for pedestrian tracking.

\end{itemize}

\section{Related Work}

\textbf{Batch Tracking vs. Online Tracking.}  Generally speaking, we can classify multi-object tracking methods into batch mode and online mode. For methods in the batch mode, video frames from future time steps can be utilized to solve the data association problem \cite{pirsiavash2011globally,butt2013multi,milan2014continuous}. Batch methods are useful for offline video analysis applications. However, they are not applicable to problems where immediate decisions have to be made for each video frame such as in robotics and autonomous driving. In contrast, for methods in the online mode, only the previous video frames and the current video frame can be used to solve the data association problem \cite{breitenstein2011online,kim2012online,yoon2015bayesian,xiang2015learning}. In online tracking, the feature representation of the object is critical for reliable data association. While most of the previous online tracking methods use hand-crafted features, we propose to encode the feature representation of object into the memory of our recurrent autoregressive networks.

\textbf{Multi-Object Tracking with Neural Networks.} Recently, a few methods have been proposed to employ deep neural networks in multi-object tracking. \cite{lealcvprw2016} learns a Siamese network to match a pair of object detections, and uses the output from the Siamese network as a similarity score for data association. Since the network is trained to match object detections, it is not able to capture long term history of the object. \cite{milan2016online} introduces an online multi-object tracking method using a recurrent neural network, where the RNN is trained for data association of multiple objects end-to-end. The end-to-end training requires significant amount of training trajectories, which limits the tracking performance of \cite{milan2016online}. In contrast, we focus on learning a good feature representation of objects with RANs, which is used in a subsequent data association algorithm for online multi-object tracking.

\textbf{Deep Autoregressive Models.} The idea of introducing autoregressive structure into deep generative models first appeared in NADE \cite{larochelle2011neural} and DARN \cite{gregor2013deep}, both of which essentially used autoregressive structure to specify the conditional dependencies and designed Bayesian networks accordingly to model the distribution of fixed-dimensional random variables. However, our RAN framework is aimed to model the temporal or sequential data, which better resonates the traditional applications of autoregressive models.


\section{Our Model} \label{sec:technical}
\begin{figure*}
	\centering
	\includegraphics[width = 0.8\linewidth]{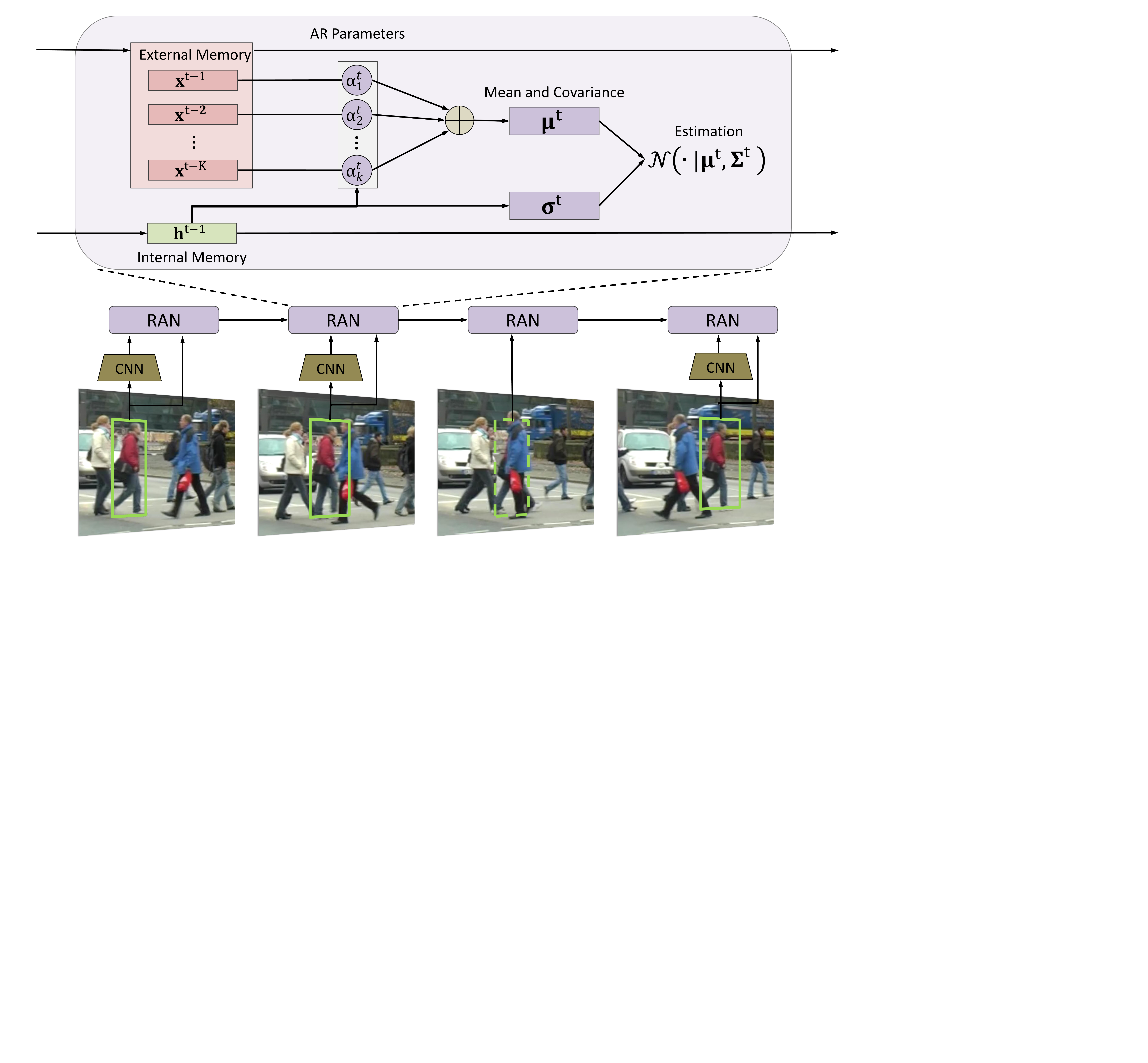}
	\caption{\textbf{Recurrent Autoregressive Network architecture.} Here we explain how an RAN is associated with a specific object trajectory. The RAN takes appearance and motion features as inputs and updates both internal and external memories based on the data association results. Here the solid boxes represent the chosen detections while the dashed box is the predicted box position from RAN. By combining internal and external memories, the RAN estimates conditional distribution of the detection in future frames as explained in Section~\ref{sec:technical}.}
	\label{fig:model}
    \vspace{-4mm}
\end{figure*}

The primary goal of our model is to memorize the characteristics of multiple object trajectories over time and use the memory for data association in a tracking-by-detection manner. Our proposed architecture (see Figure~\ref{fig:model}) couples an internal memory and an external memory for each object trajectory, as representations of appearance and motion dynamics of the object. In this section, we describe our neural network design (Section~\ref{sec:ran}), our multi-object tracking algorithm (Sections~\ref{sec:association} and ~\ref{sec:track_multiple}), and the training procedure of our method (Section~\ref{sec:training}).

\subsection{Recurrent Autoregressive Networks} \label{sec:ran}

We introduce our Recurrent Autoregressive Network (RAN) as a generative model for sequential data.

\textbf{Conditional probability distribution.} Suppose an RAN receives input vectors $\mathbf{x}^{1:t-1}$ ($\mathbf{x}^t \in \mathbb{R}^N$) through time step $1$ to $t - 1$, then it estimates the conditional probability distribution $\Pr(\mathbf{x}^{t} | \mathbf{x}^{1:t-1})$ of the incoming input vector $\mathbf{x}^{t}$. To model this conditional probability distribution, we utilize the autoregressive (AR) model which models the upcoming input $\mathbf{x}^{t}$ as a weighted sum of the previous $K$ input vectors plus a Gaussian noise:
\begin{equation}
\mathbf{x}^t = \sum_{k=1}^K \alpha_k^t \mathbf{x}^{t-k} + \boldsymbol\varepsilon^t,
\end{equation}
where $\boldsymbol\alpha^t = (\alpha_1^t, \ldots, \alpha_K^t)$ are the parameters of the AR model, and $\boldsymbol\varepsilon^t =  (\varepsilon_1^t, \ldots, \varepsilon_N^t)$ is the Gaussian white noise. Suppose $\varepsilon_j^t \sim \mathcal{N}(0, (\sigma_j^{t})^2)$ for $j = 1, \ldots, N$ is independently drew from a normal distribution with mean zero and standard deviation $\sigma_j^{t}$, then
\begin{equation} \label{eq:conditional}
\Pr(\mathbf{x}^{t} | \mathbf{x}^{1:t-1}) = \mathcal{N}(\mathbf{x}^{t} | \boldsymbol\mu^t, \boldsymbol\Sigma^t),
\end{equation}
with predicted mean and variance matrix
\begin{align}
\boldsymbol\mu^t &= \sum_{k=1}^K \alpha_k^t \mathbf{x}^{t-k}, \\
\boldsymbol\Sigma^t &=\text{diag}((\boldsymbol\sigma^t)^2) =   \text{diag}\left((\sigma_1^{t})^2, \ldots, (\sigma_N^{t})^2\right).
\end{align}
Notice that in the traditional AR models, the parameters $\boldsymbol\alpha$ and the standard deviations $\boldsymbol\sigma$ are usually optimized in a training phase and keep fixed during test time. While in our RAN model, we allow both variables to change over time. This enables us to adjust our estimation strategies according to newly received inputs when the characteristics of the sequence change.

\textbf{Recurrent parameter estimation.} In order to capture the long-term trend of the sequence and estimate $\boldsymbol\alpha^t$ and $\boldsymbol\sigma^t$ sequentially, we embed the Gate Recurrent Unit (GRU) \cite{cho2014properties} into our RAN model. GRU maintains and updates a hidden state $\mathbf{h} \in \mathbb{R}^d$ as the internal memory in time, where $d$ is the dimension of the hidden state. Given the hidden state $\mathbf{h}^{t-1}$ at time step $t-1$, the parameters of the AR model are estimated by applying a mapping function $f(\cdot)$, which in our case is a fully-connected layer (i.e. linear transformation), followed by specific element-wise transformations:
\begin{align}
(\hat{\boldsymbol\alpha}^t, \hat{\boldsymbol\sigma}^t) &= f(\mathbf{h}^{t-1}), \label{eq:prediction} \\
\alpha_k^{t} &= \frac{\exp(\hat{\alpha}_k^{t})}{\sum_{k'=1}^K, \exp(\hat{\alpha}_{k'}^{t})}, &k = 1, \ldots, K, \label{eq:alpha} \\
\sigma_j^{t} &= \exp(\hat{\sigma}_j^{t}), &j = 1, \ldots, N. \label{eq:sigma}
\end{align}
Equation~\eqref{eq:alpha} applies the softmax function on top of the fully connection mapping as a regularization to guarantee that the parameters $\alpha_k^t, i = 1, \ldots, K$ sum to one. Equation~\eqref{eq:sigma} guarantees that the standard deviations are positive. Up to now, we are able to estimate the conditional probability distribution $\Pr(\mathbf{x}^{t} | \mathbf{x}^{1:t-1})$ which will be used in our multi-object tracking algorithm.

\textbf{Hidden state update.} At time step $t$, the GRU takes an input $\mathbf{x}^t \in \mathbb{R}^N$ with dimension $n$ and the hidden state from the previous time step $\mathbf{h}^{t-1}$, and then generates a new hidden state $\mathbf{h}^t$  for time $t$ according to the following rules:
\begin{align}
\mathbf{z}^t &= g( \mathbf{W}_z  \mathbf{x}^t + \mathbf{U}_z \mathbf{h}^{t-1})  \label{eq:gru_z} \\
\mathbf{r}_t &= g( \mathbf{W}_r  \mathbf{x}^t + \mathbf{U}_r \mathbf{h}^{t-1}) \label{eq:gru_r} \\
\mathbf{\tilde{h}}^t &= \tanh( \mathbf{W} \mathbf{x}^t + \mathbf{U}  (r^t \odot \mathbf{h}^{t-1})) \label{eq:gru_h} \\
\mathbf{h}^t &= (1 - \mathbf{z}^t) \odot \mathbf{h}^{t-1} + \mathbf{z}^t \odot \mathbf{\tilde{h}}^t, \label{eq:gru}
\end{align}
where $g(\cdot)$ denotes the logistic sigmoid function, $\tanh(\cdot)$ denotes the hyperbolic tangent function and $\odot $ denotes element-wise multiplication. The GRU is parameterized by matrices $ \mathbf{W}_z$, $ \mathbf{W}_r$ and $ \mathbf{W}$ with dimension $d \times N$ and matrices $ \mathbf{U}_z$, $ \mathbf{U}_r$ and $ \mathbf{U}$ with dimension $d \times d$. $\mathbf{z}_t$ in Equation~\eqref{eq:gru_z} is regarded as the update gate, which decides the degree of update the GRU performs. $\mathbf{r}_t$ in Equation~\eqref{eq:gru_r} is called the reset gate, which decides how much information from the previous time step should be forgotten. $\mathbf{\tilde{h}}_t$ in Equation~\eqref{eq:gru_h} is known as the candidate activation, which is used to compute the new hidden state in Equation~\eqref{eq:gru}. In training, the parameters of the GRU are learned in order to update its hidden state in an appropriate way.

\textbf{External and internal memories.} We can see that our RANs maintain two types of memories about the tracked object at time $t$. The first type is an external memory that consists of $K$ input vectors in the previous time steps $\mathcal{E}^t = \{ \mathbf{x}^{t-1}, \ldots, \mathbf{x}^{t-K} \}$, which can be considered to be the templates of the object as in online single object tracking methods. The second type is an internal memory which is represented by the hidden state of RNNs $\mathbf{h}^{t-1}$. The RNN hidden state encodes information about how these templates should be combined in order to predict the probability distribution of the next input.

\subsection{Data Association with RANs} \label{sec:association}

Given a set of detections at time $t$, the data association problem for each object is to decide which detection (or none of them) the object should be associated to. We handle the data association problem using both the appearance information and the motion dynamics of the object. Let the detections be indexed by $i = 1, \ldots, M$, we compute the motion dynamic features from the bounding boxes as $\{\mathbf{b}_i^t\}_{i=1}^M$, and extract the appearance features $\{\boldsymbol\phi_i^t\}_{i=1}^M$ from the image patches of the detections. For each object $l$, let the bounding boxes and appearance features of the previous chosen detections be $\mathbf{b}_l^{1:t-1}$ and $\boldsymbol\phi_l^{1:t-1}$ respectively. Then we define the association score $s_{i | l}^t$ between detection $i$ and object $l$ as the conditional probability of $\mathbf{b}_i^t$ and $\boldsymbol\phi_i^t$ given $\mathbf{b}_l^{1:t-1}$ and $\boldsymbol\phi_l^{1:t-1}$: 
\begin{align}\label{eqn:tracking_score}
s_{i | l}^t &= \Pr(\boldsymbol\phi_i^t, \mathbf{b}_i^t | \boldsymbol\phi_l^{1:t-1}, \mathbf{b}_l^{1:t-1}) \nonumber \\
&= \Pr(\boldsymbol\phi_i^t | \boldsymbol\phi_l^{1:t-1}) \Pr(\mathbf{b}_i^t | \mathbf{b}_l^{1:t-1}) \\
&= \mathcal{N}(\boldsymbol\phi^t | \boldsymbol\mu_{\phi}^t, \boldsymbol\Sigma_{\phi}^t) \mathcal{N}(\mathbf{b}^t | \boldsymbol\mu_{b}^t, \boldsymbol\Sigma_{b}^t),
\end{align}
where both $\Pr(\boldsymbol\phi_i^t | \boldsymbol\phi_l^{1:t-1})$ and $\Pr(\mathbf{b}_i^t | \mathbf{b}_l^{1:t-1})$ are modeled with RANs as in Equation~\eqref{eq:conditional}. Here we assume the independence of bounding boxes and appearance features. After computing the association scores for all the detections, object $l$ is associated to the detection $i^*$ with maximum score if $s_{i^*|l}^t$ is larger than a predefined threshold. Otherwise, object $l$ is not associated to any detection at time $t$, which is marked as lost.

Appearance and motion are two very different modalities in terms of dimensions and updates. Therefore, we use two sibling internal memories $\mathbf{h}_{\phi}^{t}$ and $\mathbf{h}_{b}^{t}$ with two sibling external memories $\mathcal{E}_{\phi}^{t}$ and $\mathcal{E}_{b}^{t}$ for appearance and motion respectively. During tracking, when a new detection is associated, the two types of memories are updated simultaneously. $\boldsymbol\mu_{\phi}^t$, $\boldsymbol\sigma_{\phi}^t$ and $\boldsymbol\mu_{b}^t$, $\boldsymbol\sigma_{b}^t$ are derived respectively as in Equation~\eqref{eq:prediction} \eqref{eq:alpha} \eqref{eq:sigma} and then combined together to compute the association score as in \eqref{eqn:tracking_score}. When the target is lost, the $\mathbf{h}_{b}^{t}$ and $\mathcal{E}_{b}^{t}$ are updated with the predicted motion $\boldsymbol\mu_b^t$, while the appearance counterparts remain the same. 

\subsection{Tracking Multiple Objects}\label{sec:track_multiple}


When tracking multiple objects, we assemble multiple RANs, where each RAN corresponds to an object. These RANs are used to compute association scores between tracked objects and detections. Following the practice of \cite{xiang2015learning, choi2015near}, we implemented a multi-object tracking framework using bipartite matching as described in Algorithm \ref{algo:multi_object_tracking.}. For notational simplicity, here we bind all object trajectories with their corresponding internal and external memories denoted as $\mathcal{T}$. At each time step, after detections and feature extractions are finished, we first generate a set of candidate pairs $\mathcal{P}$ of trajectories and detections. For the concern of computation time, we only consider detection bounding boxes within a distance to the last detected bounding box in the trajectory. Then the association scores $s_{i|l}^t$ for each $(i, l) \in \mathcal{P}$ are computed as explained in section \ref{sec:association}. And then we associate object trajectories with detections and initialize new trajectories with unassociated detections. At the end of each time step, we update all the trajectories in $\mathcal{A}$ with their internal and external memories.

The initialization and termination of each trajectory are handled at the end of each time step. The unassociated detections in $\mathcal{U}$ are used to initialize new object trajectories. As in \cite{xiang2015learning}, we terminate a trajectory when it has been lost for more than $t_{terminate} = 20$ time steps. To reduce the number of false positive detections, standard non-maximum suppression and thresholding of detection scores are applied as in previous works. In most cases a trajectory initialized by a false detection will fail to associate to detections in the following time steps. Therefore such a trajectory will usually be marked as lost later on and eventually be terminated.

\begin{algorithm}
\caption{Our multi-object tracking approach. }\label{algo:multi_object_tracking.}
\begin{algorithmic}[1]
\Require video frames $V = I_1, ..., I_T$
\Ensure trajectories $\mathcal{T}$
\For {$t = 1, ..., T$}
\State Detect boxes $\mathcal{D}^t$ with input image $I_t$
\State Extract motion features $\{\mathbf{b}_i^t\}_{i=1}^M$
\State Extract appearance features $\{\mathbf{\phi}_i^t\}_{i=1}^M$ on $I_t$
\State Generate candidate pairs $\mathcal{P}$
\ForAll {$(i, l) \in \mathcal{P}$}
\State Compute $s_{i|l}^t$ using RANs as in Section~\ref{sec:association}
\EndFor
\State Associate $\mathcal{T}$ with $\mathcal{D}^t$ using $\{s_{i|l}^t\}_{(i, l) \in \mathcal{P}}$ as in \cite{xiang2015learning}
\State Terminate trajectories lost for more than $t_{terminate}$ steps. 
\State Initialize new trajectories with unassociated detections
\State Update $\mathcal{T}$
\EndFor
\end{algorithmic}
\end{algorithm}

\subsection{Training RANs for Tracking}\label{sec:training}

During training, our goal is to learn the RAN parameters to discriminate ground truth associations and false associations. We formulate the training procedure as a maximum likelihood estimation problem for the conditional probability distribution of the RAN. In each training iteration, we sample a batch of object trajectories from the training videos. For each trajectory $l$, instead of using the ground truth bounding boxes, we sample bounding boxes $\mathbf{b}_l^{1:T}$ among the detections whose Intersection of Unions (IOUs) with a ground truth bounding box are larger than 0.5. Then, we extract the corresponding appearance features as $\boldsymbol\phi_l^{1:T}$. We feed the features into the RAN at each time step and estimate the conditional probability distribution as in Equation~\eqref{eqn:tracking_score}. We skip the time steps when the ground truth object is invisible or lost. Finally, the training loss $\mathcal{L}$ is defined as the sum of the negative log likelihood of $\mathbf{b}_l^{1:T}$ and $\boldsymbol\phi_l^{1:T}$:
\begin{equation}
\mathcal{L} = - \sum_{l} \sum_{t} \log\Pr(\boldsymbol\phi_l^{t+1}, \mathbf{b}_l^{t+1} | \boldsymbol\phi_l^{1:t}, \mathbf{b}_l^{1:t}).
\end{equation}
This loss function encourages the RANs to predict higher probability densities around the correctly associated detections than other detections in the feature space.

\subsection{Implementation Details}\label{sec:implementaion}

\textbf{Feature Extraction.}
For the appearance features, we use the fc8 layer of an inception network \cite{Xiao_2016_CVPR} pretrained on person re-identification datasets. The feature extraction takes 0.45 second in average for each frame during test time. The extracted appearance feature vector is 256-dimensional. For motion features, we simply use a 4-dimensional vector, where the first 2 dimensions are the x-y coordinates of the detection center relative to the previous box in the trajectory, and the last 2 dimensions are the width and height of the detection. In Section~\ref{sec:experiments}, we compare the tracking performance by directly using the features with that using our RANs.

\textbf{Network Architecture.}
We choose to use 128-dimensional internal memories for appearance and 32-dimensional internal memories for motion. For external memories, the dimensions are the same with the features. The time spans of the external memories are both 10. In Section~\ref{sec:experiments}, we discuss about the influence of varying time spans.

\textbf{Optimization.}
Our training batch consists of 64 object trajectories from multiple training videos. The trajectories are randomly subsampled and temporally cropped across time for data augmentation. We use Adam \cite{Kingma2014AdamAM} for optimization with a learning rate of $1\times10^-3$ and set $\beta_1 = 0.9$, $\beta_2 = 0.99$. RnnDrop \cite{Moon2015RNNDROPAN, Gal2016ATG} are used in recurrent layers to prevent overfitting.

\begin{table} \setlength{\tabcolsep}{2pt}
\small
\centering
\begin{tabular}{l | c c c c c c}
Model & MOTA$(\uparrow)$ & MT$(\uparrow)$ & ML$(\downarrow)$ & FP$(\downarrow)$ & FN$(\downarrow)$ & IDS$(\downarrow)$ \\
\hline
\hline
A-GRU &43.3 &21.4\% &34.2\% & \textbf{1,482} &11,501 &107 \\
A-AVE &67.8 &50.9\% &15.8\% &2,156 &5,135 &138 \\
A-TIV &68.9 &51.0\% &15.8\% &2,170 &4,884 &108 \\
A-RAN & \textbf{69.9} & \textbf{53.8\%} & \textbf{12.4\%} & 2,189 & \textbf{4,684} & \textbf{80} \\
\hline
M-GRU &56.7 &48.7\% &19.7\% &2,481 &7,419 &\textbf{108} \\
M-AVE &68.5 &47.4\% &14.9\% &2,078 &5,043 &158 \\
M-TIV &68.6 &47.9\% &15.4\% & \textbf{2,054} &5,036 &149 \\
M-RAN & \textbf{68.9} & \textbf{54.3\%} & \textbf{14.1\%} & 2,752 & \textbf{4,309} &118 \\
\hline
(A+M)-GRU &57.7  &42.3\% &15.4\% &3,323 &6,362 &85 \\
(A+M)-AVE &68.6  &50.4\% &15.8\% &2,126 &4,991 &142 \\
(A+M)-TIV &69.3 &50.1\% &16.2\% &\textbf{1,992} &4,981 &109 \\
(A+M)-RAN &\textbf{70.7}  & \textbf{55.5\%} & \textbf{14.1\%} &2,123 &\textbf{4,567} &\textbf{77} \\
\hline
\end{tabular}
\caption{Analysis of the RAN tracking framework on the validation set of the 2DMOT2015 dataset.}
\label{tab:components}
\vspace{-2mm}
\end{table}

\begin{table} \setlength{\tabcolsep}{4pt}
	\small
	\centering
	\begin{tabular}{l | c c c c c c c}
		Span & MOTA$(\uparrow)$ & MT$(\uparrow)$ & ML$(\downarrow)$ & FP$(\downarrow)$ & FN$(\downarrow)$ & IDS$(\downarrow)$ \\
		\hline
		\hline
		1 &69.7 &\textbf{58.1}\% &15.4\% &2,133 &4,752 &120 \\
		2 &70.2 &54.7\% &15.4\% &2,111 &46,78 &100 \\
		3 &70.4 &56.0\% &14.1\% &2,117 &46,31 &86 \\
		4 &70.4 &55.6\% &13.7\% &2,110 &4,645 &84 \\
		5 &70.6 &56.4\% &13.6\% &2,155 &4,540 &83 \\
		6 &70.6 &55.1\% &\textbf{10.8\%} &2,066 &4,645 &84 \\
		7 &70.6 &56.4\% &13.2\% &2,123 &4,577 &85 \\
		8 &70.6 &55.1\% &12.8\% &2,112 &4,584 &81 \\
		9 &\textbf{70.7} &55.1\% &14.1\% &2,116 &4,568 &\textbf{77} \\
		10 &\textbf{70.7}  &55.5\% &14.1\% &2,123 &\textbf{4,567} &\textbf{77} \\
		11 &70.5 &54.7\% &14.1\% &2,114 &4,610 &79 \\
		12 &70.6 &55.6\% &14.1\% &\textbf{2,058} &4,575 &\textbf{77} \\
		\hline
	\end{tabular}
	\caption{MOT performance on the validation set of the MOT2015 dataset according to different time spans of the external memory.}
	\label{tab:time_span}
    \vspace{-8mm}
\end{table}

\begin{table*}
	\small
	\centering{
		\begin{tabular}{l|c|c|c|c|c|c|c|c|c}
			\toprule  Method & Mode & MOTA$(\uparrow)$ & MOTP$(\uparrow)$ & MT$(\uparrow)$  & ML$(\downarrow)$  & FP$(\downarrow)$ & FN$(\downarrow)$ & IDS$(\downarrow)$ & Frag$(\downarrow)$\\
            \midrule
            CNNTCM \cite{Wang2016JointLO} &Batch &29.6 &71.8 &11.2\% &44.0\% &7,786 &34,733 &712 &943\\
            MHT\_DAM \cite{Kim2015MultipleHT} &Batch &32.4 &71.8 &\textbf{16.0\%} &\textbf{43.8\%} &9,064 &\textbf{32,060} &\textbf{435} &826\\
            NOMT \cite{choi2015near} &Batch &\textbf{33.7} &\textbf{71.9} &12.2\% &44.0\% &\textbf{7,762} &32,547 &442 &\textbf{823}\\
        	\midrule
            SCEA \cite{Yoon2016OnlineMT} &Online &29.1 &71.1 &8.9\% &47.3\% &\textbf{6,060} &36,912 &604 &\textbf{1,182}\\
            MDP \cite{xiang2015learning} &Online &30.3 & 71.3 & 13.0\% & 38.4\% &9,717 & 32,422 &680 &1,500 \\
            AMIR15 \cite{sadeghian2017tracking} & Online & \textbf{37.6} & \textbf{71.7} & \textbf{15.8\%} & \textbf{26.8\%} & 7,933 & \textbf{29,397} & 1,026 & 2,024 \\
           	Our Model (RAN) &Online & 35.1 &70.9 & 13.0\% &42.3\% &6,771 &32,717 &\textbf{381} &1,523 \\
            \bottomrule
		\end{tabular}
		\caption{Tracking performance on the \textbf{2DMOT2015} dataset with \textbf{DPM detections}.}
		\label{table:2dmot2015_test_public}
		\vspace{-4mm}
	}
\end{table*}

\begin{table*}
	\small
	\centering{
		\begin{tabular}{l|c|c|c|c|c|c|c|c|c}
			\toprule  Method & Mode & MOTA$(\uparrow)$ & MOTP$(\uparrow)$ & MT$(\uparrow)$  & ML$(\downarrow)$  & FP$(\downarrow)$ & FN$(\downarrow)$ & IDS$(\downarrow)$ & Frag$(\downarrow)$\\
        	\midrule
            JMC \cite{Tang2016MultipersonTB} &Batch &46.3 &75.7 &15.5\% &\textbf{39.7\%} & 6,373 &90,914 &657 &1,114\\
            NOMT \cite{choi2015near} &Batch & 46.4 & 76.6 &\textbf{18.3\%} &41.4\% &9,753 &\textbf{87,565} &\textbf{359} &\textbf{504} \\
            NLLMPa \cite{levinkovjoint} &Batch &\textbf{47.6} &\textbf{78.5} &17.0\% &40.4\% & \textbf{5,844} &89,093 &629 &768 \\
            \midrule
            EAMTT \cite{eamtt2016} &Online &38.8 &\textbf{75.1} &7.9\% &49.1\% &8,114 &102,452 &965 &1,657 \\
            oICF \cite{Kieritz2016OnlineMT} &Online &43.2 &74.3 &11.3\% &48.5\% &\textbf{6,651} &96,515 &\textbf{381} &\textbf{1,404}\\
            Our Model (RAN) &Online &\textbf{45.9} &74.8 &\textbf{13.2\%} &\textbf{41.9\%} &6,871 &\textbf{91,173} &648 &1,992 \\
            \bottomrule
		\end{tabular}
		\caption{Tracking performance on the \textbf{MOT16} dataset with \textbf{DPM detections}.}
		\label{table:mot16_test_public}
		\vspace{-4mm}
	}
\end{table*}

\section{Experiments} \label{sec:experiments}

\textbf{Datasets.} We use the 2DMOT2015 \cite{MOTChallenge2015} and MOT16 \cite{Milan2016MOT16AB} datasets in our experiments. The two datasets are composed of 14 and 22 pedestrian tracking videos, with 1,221 and 1,276 object trajectories respectively. In each dataset, the videos are aggregated from multiple multi-object tracking benchmarks and equally divided into training and testing sets. We split the training set of 2DMOT2015 as 5 training videos and 6 validation videos as suggested by \cite{xiang2015learning}. 

\textbf{Evaluation Metrics.} We use multiple metrics suggested by the MOT Benchmark to evaluate the multiple object tracking performance. These are Multiple Object Tracking Accuracy (MOTA) \cite{keni2008evaluating}, Multiple Object Tracking Precision (MOTP) \cite{keni2008evaluating}, the number of ID Switches (IDS), the percentage of Mostly Track targets (MT), the percentage of Mostly Lost targets (ML), the total number of False Positives (FP), the total number False Negatives (FN), the total number of times a trajectory is Fragmented (Frag) and the frame rate of the tracking phase (Hz). Among these, MOTA and IDS are the two metrics that most directly depict the quality of tracking and association.


\textbf{Detections.} In order to compare the tracking performance on the MOT leaderboard, we run our RAN with both public detections provided by the MOT benchmark and Faster-RCNN detections \cite{ren2015faster} used by the current leading tracking algorithm. The public detections are computed by the DPM V5 detector \cite{MOTChallenge2015}. The Faster-RCNN detections are from \cite{xiang2015learning, Yu2016POIMO} using Faster-RCNN with the VGG16 network architecture \cite{simonyan2014very}. All of our analysis on the validation set use Faster-RCNN detections.

\subsection{Analyze Internal and External Memories}

In this section, we analyze the effectiveness of our RAN model design by comparing with different control settings and baseline models. In Table~\ref{tab:components}, we compare the performance of three modality settings including using only appearance information (A), using only motion dynamics (M), and using both modalities (A + M). For each modality setting, we evaluate the baseline models explained below:
\begin{itemize}
\item Gated Recurrent Unit (GRU): Instead of estimating the AR parameters and processing the external memory, we use a GRU model to directly predict the mean feature vectors along with the standard deviation variables of a multivariate Gaussian distribution. 
\item Average (AVE): We only keep the temporal sliding window of each object trajectory. The features are directly predicted by averaging all the valid features in the sliding window. The standard deviation terms are trained as time invariant variables in this case.
\item Time Invariant AR (TIV): We train a traditional AR model with time invariant parameters and standard deviations on the training videos.
\item Our full model (RAN): Both internal and external memories are used as explained in Section~\ref{sec:technical}.
\end{itemize}
AVE and TIV estimate the conditional probability distribution of future detections by using the external memories only, while the GRU baseline only uses the internal memories in the recurrent cells. To have a fair comparison of the representation capability, the update rules, the recurrent cell dimensions and the time span of the sliding window in the baselines are chosen to be the same as in our full model. The association threshold for each trained model is chosen by an automatic random search. 


\begin{table*}
	\small
	\centering{
		\begin{tabular}{l|c|c|c|c|c|c|c|c|c}
			\toprule  Method & Mode & MOTA$(\uparrow)$ & MOTP$(\uparrow)$ & MT$(\uparrow)$  & ML$(\downarrow)$  & FP$(\downarrow)$ & FN$(\downarrow)$ & IDS$(\downarrow)$ & Frag$(\downarrow)$\\
			\midrule
			TSML + CDE \cite{DBLP:journals/corr/WangWCW15} & Batch & 49.1 & 74.3 & 30.4\% & 26.4\% & \textbf{5,204} & 25,460 & 637 & 1,034 \\
			NOMT + SDP \cite{choi2015near} & Batch & \textbf{55.5} & \textbf{76.6} & \textbf{39.0\%} & \textbf{25.8\%} & 5,594 & \textbf{21,322} & \textbf{427} & \textbf{701} \\
			\midrule
			SORT \cite{DBLP:journals/corr/BewleyGORU16} & Online & 33.4 & 72.1 & 11.7\% & 30.9\% & \textbf{7,318} & 32,615 & 1,001 & 1,764\\
			MDP + SubCNN \cite{xiang2015learning} & Online & 47.5 & 74.2 & 30.0\% & 18.6\% & 8,631 & 22,969 & 628 & 1,370 \\
			EAMTT \cite{eamtt2016} & Online & 53.0 & \textbf{75.3} & 35.9\% & 19.6\% & 7,538 & 20,590 & 776 & \textbf{1,269} \\
			Our Model (RAN) & Online & \textbf{56.5} & 73.0 & \textbf{45.1\%} & \textbf{14.6\%} & 9,386 & \textbf{16,921} & \textbf{428} & 1,364 \\
			\bottomrule
		\end{tabular}
		\caption{Multi-object tracking performance on the test set of the \textbf{2DMOT2015} dataset with \textbf{Faster-RCNN detections}.}
		\vspace{-2mm}
		\label{table:2dmot2015_test}
	}
\end{table*}

 \begin{table*}
 	\small
 	\centering{
 		\begin{tabular}{l|c|c|c|c|c|c|c|c|c}
 			\toprule  Method & Mode & MOTA$(\uparrow)$ & MOTP$(\uparrow)$ & MT$(\uparrow)$  & ML$(\downarrow)$  & FP$(\downarrow)$ & FN$(\downarrow)$ & IDS$(\downarrow)$ & Frag$(\downarrow)$\\
 			\midrule
 			NOMT + SDP \cite{choi2015near} & Batch & 62.2 & \textbf{79.6} & \textbf{32.5\%} & 31.1\% & \textbf{5,119} & 63,352 & \textbf{406} & \textbf{642}\\            
 			MCMOT\_HDM \cite{lee2016multi} & Batch & \textbf{62.4} & 78.3 & 31.5\% & \textbf{24.2\%} & 9,855 & \textbf{57,257} & 1,394 & 1,318 \\
 			\midrule
 			EAMTT \cite{eamtt2016} & Online & 52.5 & 78.8 & 19.9\% & 34.9\% & \textbf{4,407} & 81,223 & 910 & 1,321 \\            
 			SORT \cite{DBLP:journals/corr/BewleyGORU16} & Online & 59.8 &	\textbf{79.6} &	25.4\% & 22.7\% &	8,698 & 63,245 & 1,423 & 1,835 \\
 			POI \cite{Yu2016POIMO} & Online & \textbf{66.1} & 79.5 & 34.0\% & \textbf{20.8\%} & 5,061 & 55,914 & 805 & 3,093 \\
 			Our Model (RAN) & Online & 63.0 & 78.8 & \textbf{39.9\%} & 22.1\% & 13,663& \textbf{53,248} & \textbf{482} & \textbf{1,251} \\
 			\bottomrule
 		\end{tabular}
 		\caption{Multi-object tracking performance on the test set of the \textbf{MOT16} dataset with \textbf{Faster-RCNN detections}.}
 		\label{table:mot16_test}
 		\vspace{-4mm}
 	}
 \end{table*}

\begin{figure*}
	\centering
	\includegraphics[height = 0.5\linewidth,width = 1.0\linewidth]{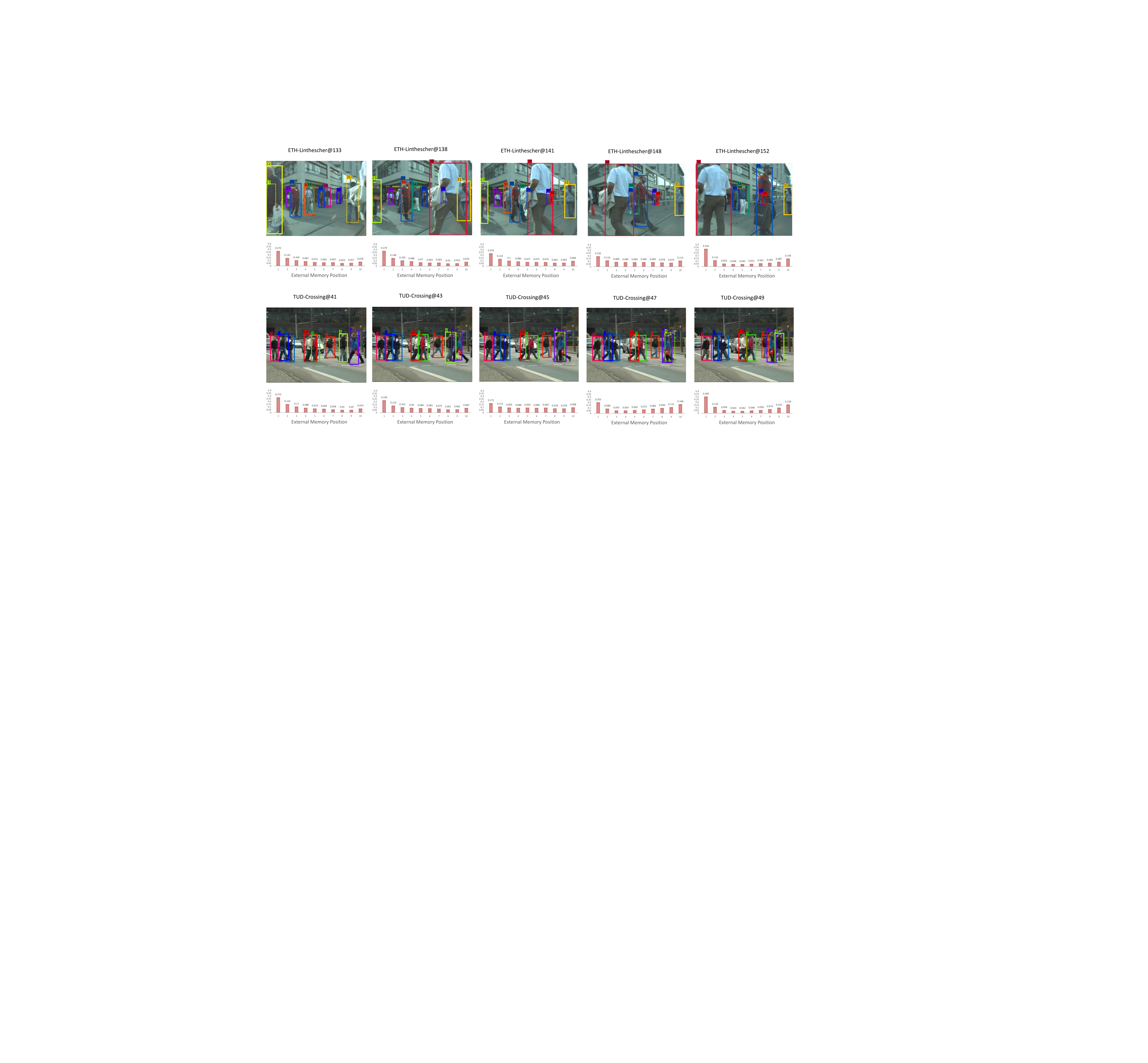}
	\caption{\textbf{Visualization of the tracking results of multiple pedestrians with RANs in highly occluded scenes and the predicted RAN parameters over time.} On the top row, we visualize the the 133, 138, 141, 148 and 152 frames in the ETH-Linthescher video. The predicted RAN parameters of object 25 for appearance features are visualized in histograms over time. In the bottom row, we visualize the the 41, 43, 45, 47, 49 frames in the TUD-Crossing video. The predicted RAN parameters of object 8 for appearance features are plotted. }
	\label{fig:qualitative}
\end{figure*}

The tracking performance of the baselines and model variants are summarized in Table~\ref{tab:components}. Within each modality setting, the RAN models obtain the best MOTA and IDS while the GRU baselines have the worst results. The performance difference is most obvious for the appearance modality, since the high dimensional CNN appearance features require more representation power in the neural network memories. Comparing the three models with external memories, using a set of trainable parameters has comparable or better performance than directly averaging the templates. For TIVs, usually the weight of the most recent template in the sliding window is highest the weights decay exponentially for previous time steps. For RANs, the predicted weights also decay in time in most cases, but their values vary when occlusions and noisy detections happen.


Among the three modalities, the best performance is achieved by using both appearance and motion. Between using only appearance or motion, the former achieves fewer IDS for each model design, since in most cases appearance is a more discriminative clue for different objects than bounding box motions. For the GRU baseline, using motions obtain higher MOTA than using appearance. When using motion only in our current setting, TIV and RAN can hardly do better than AVE, which is equivalent to a naive constant velocity motion prior widely used in previous tracking frameworks \cite{xiang2015learning, choi2015near}. A obtains much fewer ID switches and higher MOTA than M. This is consistent with our observations that both appearance and motion affects the tracking performance, while the appearance information is more discriminative when tracking targets have overlaps and occlusions with each other. 

\subsection{Analyze the Time Span of the External Memory}

For our full RAN model framework, we vary the time span of the external memory and analyze its influence on the tracking performance. Previous works \cite{xiang2015learning} show that using longer sliding window is able to capture richer history information. In \cite{xiang2015learning}, the tracking performance fluctuates as the time span being extended and the peak is reached at certain length. In Table~\ref{tab:time_span}, we increase the time span from 1 time step to 12 time steps. The observed MOTA and IDS have a leap at the beginning, then constantly increase with small fluctuations until converging after 9 steps. This demonstrates the robustness of our RAN tracking framework in terms of architecture variants and trajectory length. RAN models with longer time span are harder to train since more GPU memory will be required during training. While at test time the majority time is spent on forwarding the CNN, while the time span has minor influence on the runtime.

\subsection{Evaluation on Test Set}


In order to evaluate our method on the 2DMOT2015 and MOT16 test set, we submit our results to the MOT Benchmark evaluation server and compare with the state-of-the-art multi-object tracking methods. As shown in Table~\ref{table:2dmot2015_test_public} and Table~\ref{table:mot16_test_public}, our RAN model achieves the competitive MOTA and IDS among all online methods using public detections. Table~\ref{table:2dmot2015_test} shows that we achieved the new state-of-the-art performance in terms of the MOTA and IDS comparing with other online methods on 2DMOT2015. Our method even outperforms the batched methods in term of MT, ML and FN. On MOT16, our method achieves the best MT, FN and IDS among all online methods. As shown in Table~\ref{table:mot16_test}, our MOTA performance reaches the state-of-the-art results \cite{Yu2016POIMO}. For per video performance, our RAN tracking framework outperforms \cite{Yu2016POIMO} in terms of MOTA in 5 of the 7 testing videos and in terms of IDS for all the testing videos.

\subsection{Qualitative Analysis}

In this section, we show qualitative results of our RAN tracking framework in challenging scenes and demonstrate how our RAN predicts parameters changes over time. In Figure~\ref{fig:qualitative}, we choose two highly crowded and occluded videos from the testing videos and show the tracking results and the predicted RAN parameters.

The top row shows selected frames from the ETH-Linthescher video, where complicated occlusions continuously happen when a group of people walk across a business street. The predicted parameters has an exponential decay across time with a rising tail at the end. When the object keep being visible in the video, the parameters rely more on the last feature. At time step 141 when the occlusion is about to happen and the chosen detection box becomes noisy, the updated parameters lean more towards memorized features in previous time steps in the external memory. When the object reappears in the video, the RAN chose to use a protective estimation strategy which is closer to averaging features in the external memory. Later the updated parameters choose to trust more on the most recent frame when the appearance became more stable.

In the bottom row, we show another example from the TUD-Crossing video. This video is shot at a crossing where multiple people are walking opposite directions on a crosswalk. This video is challenging because all the pedestrians who are walking parallel in front of the camera have similar scales and velocities with each other, so the tracking algorithm needs to rely on objects' appearance rather than bounding box motions to associate the objects. When object 8 enters the scene without occlusions, the predicted parameters are in a similar distribution with the previous example. As object 8 starting to overlap with object 4, the parameters lean to the previous frames in time step 43 and 45. Starting from time step 47, the two overlapped objects move away from each other. And the updated parameters gradually get back to the normal case.

\section{Conclusion}


In this work, we propose a novel recurrent autoregressive network for online multi-object tracking. Our RAN maintains an external memory and an internal memory in order to capture the history and the characteristics of an object during tracking. In order to track multiple objects, we represent each object with a RAN, and solve the data association problem by computing likelihoods of object detections according to the distribution modeled by the RAN. Experiments are conducted on a commonly-used benchmark for multi-object tracking, which demonstrate the advantages of our method for online multi-object tracking. We believe the idea of RAN in combining external memories and internal memories can be useful in other sequential data modeling and video analysis tasks.

\section*{Acknowledgement}
We acknowledge the support of ONR (1196026-1-TDVWE), NISSAN (1188371-1-UDARQ) and NVIDIA.


\begin{thebibliography}{10}\itemsep=-1pt

\bibitem{DBLP:journals/corr/BewleyGORU16}
A.~Bewley, Z.~Ge, L.~Ott, F.~Ramos, and B.~Upcroft.
\newblock Simple online and realtime tracking.
\newblock {\em CoRR}, abs/1602.00763, 2016.

\bibitem{breitenstein2011online}
M.~D. Breitenstein, F.~Reichlin, B.~Leibe, E.~Koller-Meier, and L.~Van~Gool.
\newblock Online multiperson tracking-by-detection from a single, uncalibrated
  camera.
\newblock {\em TPAMI}, 33(9):1820--1833, 2011.

\bibitem{butt2013multi}
A.~A. Butt and R.~T. Collins.
\newblock Multi-target tracking by lagrangian relaxation to min-cost network
  flow.
\newblock In {\em CVPR}, pages 1846--1853, 2013.

\bibitem{Cai1998FunctionalcoeecientRM}
Z.~Cai and J.~Fan.
\newblock Functional-coeecient regression models for nonlinear time series.
\newblock 1998.

\bibitem{cho2014properties}
K.~Cho, B.~Van~Merri{\"e}nboer, D.~Bahdanau, and Y.~Bengio.
\newblock On the properties of neural machine translation: Encoder-decoder
  approaches.
\newblock {\em arXiv preprint arXiv:1409.1259}, 2014.

\bibitem{choi2015near}
W.~Choi.
\newblock Near-online multi-target tracking with aggregated local flow
  descriptor.
\newblock In {\em ICCV}, pages 3029--3037, 2015.

\bibitem{Gal2016ATG}
Y.~Gal and Z.~Ghahramani.
\newblock A theoretically grounded application of dropout in recurrent neural
  networks.
\newblock In {\em NIPS}, 2016.

\bibitem{Graves2014NeuralTM}
A.~Graves, G.~Wayne, and I.~Danihelka.
\newblock Neural turing machines.
\newblock {\em CoRR}, abs/1410.5401, 2014.

\bibitem{gregor2013deep}
K.~Gregor, I.~Danihelka, A.~Mnih, C.~Blundell, and D.~Wierstra.
\newblock Deep autoregressive networks.
\newblock {\em arXiv preprint arXiv:1310.8499}, 2013.

\bibitem{keni2008evaluating}
B.~Keni and S.~Rainer.
\newblock Evaluating multiple object tracking performance: the clear mot
  metrics.
\newblock {\em EURASIP Journal on Image and Video Processing}, 2008:1:1--1:10,
  2008.

\bibitem{Kieritz2016OnlineMT}
H.~Kieritz, S.~Becker, W.~Hubner, and M.~Arens.
\newblock Online multi-person tracking using integral channel features.
\newblock In {\em AVSS}, 2016.

\bibitem{Kim2015MultipleHT}
C.~Kim, F.~Li, A.~Ciptadi, and J.~M. Rehg.
\newblock Multiple hypothesis tracking revisited.
\newblock In {\em 2015 IEEE International Conference on Computer Vision
  (ICCV)}, 2015.

\bibitem{kim2012online}
S.~Kim, S.~Kwak, J.~Feyereisl, and B.~Han.
\newblock Online multi-target tracking by large margin structured learning.
\newblock In {\em ACCV}, pages 98--111. 2012.

\bibitem{Kingma2014AdamAM}
D.~P. Kingma and J.~Ba.
\newblock Adam: A method for stochastic optimization.
\newblock {\em CoRR}, abs/1412.6980, 2014.

\bibitem{larochelle2011neural}
H.~Larochelle and I.~Murray.
\newblock The neural autoregressive distribution estimator.
\newblock In {\em AISTATS}, volume~1, page~2, 2011.

\bibitem{MOTChallenge2015}
L.~Leal-Taix\'{e}, A.~Milan, I.~Reid, S.~Roth, and K.~Schindler.
\newblock {MOTC}hallenge 2015: {T}owards a benchmark for multi-target tracking.
\newblock {\em arXiv:1504.01942 [cs]}, 2015.

\bibitem{lealcvprw2016}
L.~Leal-Taixé, C.~Canton-Ferrer, and K.~Schindler.
\newblock Learning by tracking: siamese cnn for robust target association.
\newblock {\em Computer Vision and Pattern Recognition Conference Workshops
  (CVPR). DeepVision: Deep Learning for Computer Vision.}, 2016.

\bibitem{lee2016multi}
B.~Lee, E.~Erdenee, S.~Jin, and P.~K. Rhee.
\newblock Multi-class multi-object tracking using changing point detection.
\newblock {\em arXiv preprint arXiv:1608.08434}, 2016.

\bibitem{levinkovjoint}
E.~Levinkov, J.~Uhrig, S.~Tang, M.~Omran, E.~Insafutdinov, A.~Kirillov,
  C.~Rother, T.~Brox, B.~Schiele, and B.~Andres.
\newblock Joint graph decomposition \& node labeling: Problem, algorithms,
  applications.
\newblock In {\em CVPR}, 2017.

\bibitem{Milan2016MOT16AB}
A.~Milan, L.~Leal-Taix{\'e}, I.~D. Reid, S.~Roth, and K.~Schindler.
\newblock Mot16: A benchmark for multi-object tracking.
\newblock {\em CoRR}, abs/1603.00831, 2016.

\bibitem{milan2016online}
A.~Milan, S.~H. Rezatofighi, A.~Dick, K.~Schindler, and I.~Reid.
\newblock Online multi-target tracking using recurrent neural networks.
\newblock {\em arXiv preprint arXiv:1604.03635}, 2016.

\bibitem{Milan2017OnlineMT}
A.~Milan, S.~H. Rezatofighi, A.~R. Dick, K.~Schindler, and I.~D. Reid.
\newblock Online multi-target tracking using recurrent neural networks.
\newblock {\em CoRR}, abs/1604.03635, 2017.

\bibitem{milan2014continuous}
A.~Milan, S.~Roth, and K.~Schindler.
\newblock Continuous energy minimization for multitarget tracking.
\newblock {\em TPAMI}, 36(1):58--72, 2014.

\bibitem{Moon2015RNNDROPAN}
T.~Moon, H.~Choi, H.~Lee, and I.~Song.
\newblock Rnndrop: A novel dropout for rnns in asr.
\newblock In {\em ASRU}, 2015.

\bibitem{Ondruska2016DeepTS}
P.~Ondruska and I.~Posner.
\newblock Deep tracking: Seeing beyond seeing using recurrent neural networks.
\newblock In {\em AAAI}, 2016.

\bibitem{pirsiavash2011globally}
H.~Pirsiavash, D.~Ramanan, and C.~C. Fowlkes.
\newblock Globally-optimal greedy algorithms for tracking a variable number of
  objects.
\newblock In {\em CVPR}, pages 1201--1208, 2011.

\bibitem{ren2015faster}
S.~Ren, K.~He, R.~Girshick, and J.~Sun.
\newblock Faster {R-CNN}: Towards real-time object detection with region
  proposal networks.
\newblock In {\em Neural Information Processing Systems ({NIPS})}, 2015.

\bibitem{eamtt2016}
R.~Sanchez-Matilla, F.~Poiesi, and A.~Cavallaro.
\newblock Online multi-target tracking with strong and weak detections.
\newblock In {\em Proceedings of 2nd Workshop on Benchmarking Multi-target
  Tracking: MOTChallenge 2016}, 2016.

\bibitem{simonyan2014very}
K.~Simonyan and A.~Zisserman.
\newblock Very deep convolutional networks for large-scale image recognition.
\newblock {\em arXiv preprint arXiv:1409.1556}, 2014.

\bibitem{Sukhbaatar2015EndToEndMN}
S.~Sukhbaatar, A.~Szlam, J.~Weston, and R.~Fergus.
\newblock End-to-end memory networks.
\newblock In {\em NIPS}, 2015.

\bibitem{Tang2016MultipersonTB}
S.~Tang, B.~Andres, M.~Andriluka, and B.~Schiele.
\newblock Multi-person tracking by multicut and deep matching.
\newblock In {\em ECCV Workshops}, 2016.

\bibitem{DBLP:journals/corr/WangWCW15}
B.~Wang, G.~Wang, K.~L. Chan, and L.~Wang.
\newblock Tracklet association by online target-specific metric learning and
  coherent dynamics estimation.
\newblock {\em CoRR}, abs/1511.06654, 2015.

\bibitem{Wang2016JointLO}
B.~Wang, L.~Wang, B.~Shuai, Z.~Zuo, T.~Liu, K.~L. Chan, and G.~Wang.
\newblock Joint learning of convolutional neural networks and temporally
  constrained metrics for tracklet association.
\newblock In {\em 2016 IEEE Conference on Computer Vision and Pattern
  Recognition Workshops (CVPRW)}, 2016.

\bibitem{xiang2015learning}
Y.~Xiang, A.~Alahi, and S.~Savarese.
\newblock Learning to track: Online multi-object tracking by decision making.
\newblock In {\em International Conference on Computer Vision (ICCV)}, pages
  4705--4713, 2015.

\bibitem{Xiao_2016_CVPR}
T.~Xiao, H.~Li, W.~Ouyang, and X.~Wang.
\newblock Learning deep feature representations with domain guided dropout for
  person re-identification.
\newblock In {\em The IEEE Conference on Computer Vision and Pattern
  Recognition (CVPR)}, June 2016.

\bibitem{Xiong2016DynamicMN}
C.~Xiong, S.~Merity, and R.~Socher.
\newblock Dynamic memory networks for visual and textual question answering.
\newblock In {\em ICML}, 2016.

\bibitem{Yoon2016OnlineMT}
J.~H. Yoon, C.-R. Lee, M.-H. Yang, and K.-J. Yoon.
\newblock Online multi-object tracking via structural constraint event
  aggregation.
\newblock In {\em 2016 IEEE Conference on Computer Vision and Pattern
  Recognition (CVPR)}, 2016.

\bibitem{yoon2015bayesian}
J.~H. Yoon, M.-H. Yang, J.~Lim, and K.-J. Yoon.
\newblock Bayesian multi-object tracking using motion context from multiple
  objects.
\newblock In {\em WACV}, pages 33--40, 2015.

\bibitem{Yu2016POIMO}
F.~Yu, W.~Li, Q.~Li, Y.~Liu, X.~Shi, and J.~Yan.
\newblock Poi: Multiple object tracking with high performance detection and
  appearance feature.
\newblock {\em CoRR}, abs/1610.06136, 2016.

\end{thebibliography}

\end{document}